\documentclass{nle}
\usepackage{multirow}
\usepackage{url}
\usepackage{booktabs}
\usepackage{graphicx}
\usepackage{amsmath}

\makeatletter
\newcommand{\@BIBLABEL}{\@emptybiblabel}
\newcommand{\@emptybiblabel}[1]{}
\makeatother
\usepackage[hidelinks]{hyperref}

\usepackage{color}

\title[Learning Reporting Dynamics for Rumour Detection]{Learning Reporting Dynamics during Breaking News for Rumour Detection in Social Media}
\author[Zubiaga et al.]{Arkaitz Zubiaga, Maria Liakata, Rob Procter\\ University of Warwick, Coventry, UK\\ \texttt{a.zubiaga@warwick.ac.uk}}

\pagerange{\pageref{firstpage}--\pageref{lastpage}}
\pubyear{2016}

\begin{document}

\label{firstpage}
\maketitle

\begin{abstract}
 Breaking news leads to situations of fast-paced reporting in social media, producing all kinds of updates related to news stories, albeit with the caveat that some of those early updates tend to be rumours, i.e., information with an unverified status at the time of posting. Flagging information that is unverified can be helpful to avoid the spread of information that may turn out to be false. Detection of rumours can also feed a rumour tracking system that ultimately determines their veracity. In this paper we introduce a novel approach to rumour detection that learns from the sequential dynamics of reporting during breaking news in social media to detect rumours in new stories. Using Twitter datasets collected during five breaking news stories, we experiment with Conditional Random Fields as a sequential classifier that leverages context learnt during an event for rumour detection, which we compare with the state-of-the-art rumour detection system as well as other baselines. In contrast to existing work, our classifier does not need to observe tweets querying a piece of information to deem it a rumour, but instead we detect rumours from the tweet alone by exploiting context learnt during the event. Our classifier achieves competitive performance, beating the state-of-the-art classifier that relies on querying tweets with improved precision and recall, as well as outperforming our best baseline with nearly 40\% improvement in terms of F1 score. The scale and diversity of our experiments reinforces the generalisability of our classifier.
\end{abstract}

\section{Introduction}

The use of social media to follow news stories has become commonplace in recent years. Well-known platforms such as Twitter are increasingly being used by people to learn about the latest developments \cite{sankaranarayanan2009twitterstand}, as well as by journalists for news gathering \cite{zubiaga2013curating}. This is possible thanks to the way in which they enable users to post and share updates from anywhere and at any time, hence making it possible to get reports from users on the ground who happen to witness a newsworthy event or from users that, for some reason, appear to have access to exclusive information. However, the speed at which breaking news unfolds on social media during fast-paced events, such as terrorist attacks or riots, inevitably means that much of the information posted in the early stages of news reporting is unverified \cite{procter2013readingb}. The presence of such rumours in the stream of tweets makes it more difficult for users to distinguish verified information from rumours, and coverage of the news becomes more challenging for news practitioners.

In this work we set out to develop a rumour detection system that enables flagging of unverified posts, so that one can easily distinguish information that is unsubstantiated. A rumour detection system would ultimately warn users of the unverified status of a post, letting them know that it might later be proven false; this can be useful both to limit the diffusion of information that might turn out subsequently to be false and so reduce the risk of harm to individuals, communities and society \cite{webb2016digital}. Research in rumour detection is scarce in the scientific literature, \cite{zhao2015enquiring} being the only published work to date that addresses this issue. They introduced an approach that looks for 'enquiry tweets', i.e., tweets that query or challenge the credibility of a previous posting
to determine whether it is rumourous; a tweet is deemed to be querying if it matches one of a number of manually curated, regular expressions. While this is an ingenious approach, it has some important limitations: it is reliant on there being a human in the loop to regularly revise the list of regular expressions as these may not generalise well to new datasets; it assumes that querying posts will arise, though this may lead to low recall as not all rumourous posts will necessarily provoke queries; and it takes no account of the context surrounding the information, which we believe can be exploited to gain insight into the way that a piece of information emerges. Other work has dealt with ``rumour detection'' with what we argue is a questionable definition and which conflicts with our own \cite{zubiaga2016analysing}. These studies understand rumours as false pieces of information, and therefore define the rumour detection task as consisting of distinguishing true and false stories. In our study we adhere to the prevailing definition in the scientific literature that understands a rumour as the information that is being circulated while its veracity is yet to be confirmed \cite{allport1946analysis,difonzo2007rumor}. Consequently, we define the goal of the rumour detection task as that of identifying pieces of information that are yet to be verified, distinguishing them from non-rumours. Our work makes the following contributions within the scope of this definition of the rumour detection task:

\begin{itemize}
 \item We describe a novel methodology for the collection and annotation of Twitter datasets containing a diverse range of rumours and non-rumours. Our methodology, developed in close collaboration with journalists, consists in a bottom-up approach that enables going through a timeline of tweets associated with a breaking news story to annotate rumours that were not necessarily known \textit{a priori}. Previous work has largely focused on top-down approaches that first list a set of rumourous stories known to have been circulating, and then go through the tweets to find them, which does not make possible discovery of new rumours that have not yet been listed.
 \item To the best of our knowledge, our work is the first to perform the rumour detection task without having to observe querying tweets to identify that a piece of information is rumourous. Instead, we introduce a sequential approach based on Linear-Chain Conditional Random Fields (CRF) to learn the dynamics of information during breaking news, which enables us to classify a piece of information as a rumour or non-rumour by leveraging the context learnt as the event unfolds, and relying only on the content of a tweet to determine if is rumourous. Hence, our approach does not require a tweet to trigger querying posts to determine if it is rumourous.
 \item We investigate the performance of CRF as a sequential classifier on five Twitter datasets associated with breaking news to detect the tweets that constitute rumours. The performance of CRF is compared with its non-sequential equivalent, a Maximum Entropy classifier, as well as the state-of-the-art rumour detection approach by \cite{zhao2015enquiring} and additional baseline classifiers. Our experiments show substantial improvements with CRF's use of the sequential dynamics of reporting learnt during an event as context that enriches the content of the tweet itself. These improvements are consistent across the different events in our dataset, as well as over different phases of event reporting, including in the early stages where the sequence to be exploited is more limited.
\end{itemize}

\section{Background: Definition of Rumour}

Rumours have been studied and analysed from a range of perspectives, and within and across different disciplines \cite{donovan2007idle}. Most of the definitions given in the literature agree with that of major dictionaries such as the Oxford English Dictionary, which defines a rumour as ``a currently circulating story or report of uncertain or doubtful truth'', as well as the Merriam-Webster dictionary defining it as ``information or a story that is passed from person to person but has not been proven to be true''. Irrespective of the underlying story being ultimately proven true or false, or remaining unsubstantiated, a rumour circulates while it is yet to be verified. A number of researchers have extended the definition of rumour. For instance, \cite{difonzo2007rumor} define rumours as ``unverified and instrumentally relevant information statements in circulation that arise in contexts of ambiguity, danger or potential threat, and that function to help people make sense and manage risk''. Moreover, \cite{allport1946analysis} posit that one of the main reasons why rumours circulate is that ``the topic has importance for the individual who hears and spreads the story''. In \cite{allport1947psychology}, the authors also emphasise that ``newsworthy events are likely to breed rumors'' and that ``the amount of rumor in circulation will vary with the importance of the subject to the individuals involved times the ambiguity of the evidence pertaining to the topic at issue''.

Consistent with these definitions, we adhere here to a definition adapted to the context of breaking news, which we introduced in previous work \cite{zubiaga2016analysing}: a rumour is a ``circulating story of questionable veracity, which is apparently credible but hard to verify, and produces sufficient skepticism and/or anxiety so as to motivate finding out the actual truth''. In the context of journalism, spreading rumours can have harmful consequences for the reputation of a news organisation if they are used in reporting and later proven false, and hence being able with confidence to quickly assess whether information has not yet been verified as breaking news unfolds is crucial.

\section{Related Work}

Despite the increasing interest in analysing rumours in social media \cite{procter2013readinga,procter2013readingb,starbird2014rumors,zubiaga2015crowdsourcing,takayasu2015rumor,tolosi2016analysis,zubiaga2016analysing} and the building of tools to deal with rumours that had been previously identified \cite{seo2012identifying,takahashi2012rumor}, there has been very little work in automatic rumour detection. Some of the work in rumour detection \cite{qazvinian2011rumor,hamidian2015rumor,hamidian2016rumor} has been limited to finding rumours known \textit{a priori}. A classifier is fed with a set of predefined rumours (e.g., \textit{Obama is muslim}), which then classifies new tweets as being related to one of the known rumours or not (e.g., \textit{I think Obama is not muslim} would be about the rumour, while \textit{Obama was talking to a group of Muslims} wouldn't). An approach like this can be useful for long-standing rumours, where one wants to identify relevant tweets to track the rumours that have already been identified; one may also refer to this task as \textit{rumour tracking} rather than \textit{rumour detection}. However, this would not work for fast-paced contexts such as breaking news, where new, previously unseen rumours emerge, and one does not know \textit{a priori} the specific keywords linked to the rumour, which is yet to be detected. To deal with such situations, a classifier would need to learn generalisable patterns that will help identify new rumours during breaking stories.

To the best of our knowledge, the only work that has tackled the detection of new rumours is that by \cite{zhao2015enquiring}. Their approach builds on the assumption that rumours will provoke tweets from skeptical users who question or enquire about their veracity; the fact that a piece of information has a number of querying tweets associated with it would then imply that the information is rumourous. The authors created a manually curated list of five regular expressions (e.g., ``is (that \textbar~this \textbar~it) true''), which are used to identify querying tweets. These querying tweets are then clustered by similarity, each cluster being ultimately deemed a candidate rumour. It was not viable for the authors to evaluate by recall, but their best approach achieved 52\% and 28\% precision for two datasets. While this work builds on a sensible hypothesis and presents a clever approach to tackling the rumour detection task, we foresee three potential limitations: (1) being based on manually curated regular expressions the approach may not generalise well, (2) the hypothesis might not always apply and hence lead to low recall as, for example, certain rumours reported by reputable media are not always questioned by the general public \cite{zubiaga2016analysing}, and (3) it takes no account of the context that precedes the rumour, which can give additional insights into what is going on and how a piece of information can be rumourous in that context (e.g., the rumour that \textit{a gunman is on the loose}, when the police has not confirmed it yet, is easier to be deemed a rumour if we put it into the context of the preceding events, such as additional reports that the identity of the gunman is unknown and the reasons that motivated the shooting have not been found out). In this work, we introduce a context-aware rumour detection system that uses a sequential classifier to examine the reporting dynamics during breaking news to determine if a new piece of information constitutes a rumour.

While not strictly doing rumour detection, other researchers have worked on related tasks. For instance, there is an increasing body of work \cite{qazvinian2011rumor,liu2015real,hamidian2015rumor,hamidian2016rumor,lukasik2015classifying,zeng2016unconfirmed} looking into stance classification of tweets discussing rumours, categorising tweets as supporting, denying or questioning the rumour. The approach has been to train a classifier from a labelled set of tweets to categorise the stance observed in new tweets discussing rumours; however, these authors do not deal with non-rumours, assuming instead that the input to the classifier is already cleaned up to include only tweets related to rumours. There is also work on veracity classification both in the context of rumours and beyond \cite{sun2013detecting,cai2014rumors,liang2015rumor,liu2015real,ma2015detect,wu2015false,ma2016detecting,jin2016news}. Work on stance and veracity classification can be seen as complementary to our objectives; one could use the set of rumours detected by a rumour detection system as input to a classifier that determines stance of tweets in those rumours and/or veracity of those rumours. However, this previous step of distinguishing between out rumours and non-rumours is largely unexplored, and most work deals directly with subsequent steps.



\section{Dataset}

One of our main objectives when planning to put together a dataset of rumours and non-rumours was to develop a means to collect a diverse set of stories, which would not necessarily be known \textit{a priori} and which would include both rumours and non-rumours. We did this by emulating the scenario in which a user is following reports associated with breaking news. Seeing a timeline of tweets about the breaking news, a user would then annotate each of the tweets as being a rumour or a non-rumour. To make sure that our users had the expertise to perform this annotation, we enlisted the help of a team of journalists who are partners of our research project. Our data collection approach differs substantially from that of previous work \cite{qazvinian2011rumor,procter2013readinga,starbird2014rumors}, who first identified the rumours of interest and then collected tweets associated with those by filtering using relevant keywords. By following the latter approach of gathering rumours known \textit{a priori}, one can, for instance, search for tweets with specific keywords, e.g., for tweets posted during the 2011 England Riots, one can search for `London Eye fire', to retrieve tweets associated with the rumour that the London Eye had been set on fire. However, this requires the rumour in question to be known \textit{a priori}, and will fail to identify rumours associated with events for which specific keywords have not been previously defined. We argue that this approach will miss some of the rumours, a problem that we overcome here by having journalists sift through the timeline of tweets to discover rumours. While manual annotation of the whole stream of tweets associated with breaking news is not viable, we alleviate the task by sampling the tweets that provoked a large number of retweets and hence are likely to be of interest for reporting. This is also consistent with one of the main characteristics of rumours, which tend to generate significant levels of interest.

Our data collection approach consists of harvesting tweets from the Twitter streaming API relating to newsworthy events that could potentially prompt the initiation and propagation of rumours. Collection through the streaming API was launched straight after the journalists identified a newsworthy event likely to give rise to rumours. As soon as the journalists informed us about a newsworthy event, we set up the data collection process for that event, tracking the main hashtags and keywords pertaining to the event as a whole. Note that while launching the collection slightly after the start of the event means that we may have missed the very early tweets, we kept collecting subsequent retweets of those early tweets, making it much more likely that we would retrieve the most retweeted tweets from the very first minutes. Once we had the collection of tweets for a newsworthy event in place, we sampled the timeline of tweets to enable manual annotation (signaled by highly retweeted tweets associated with newsworthy current events). Afterwards, journalists read through the timeline to mark each of the tweets as being a rumour or not, making sure that the identification of rumours was in line with the established criteria \cite{zubiaga2015crowdsourcing}.

We followed the process above for five different newsworthy events, all of which attracted substantial interest in the media and were rife with rumours:

\begin{itemize}
 \item Ferguson unrest: citizens of Ferguson in Michigan, USA, protested after the fatal shooting of an 18-year-old African American, Michael Brown, by a white police officer on August 9, 2014.
 \item Ottawa shooting: shootings occurred on Ottawa’s Parliament Hill in Canada, resulting in the death of a Canadian soldier on October 22, 2014.
 \item Sydney siege: a gunman held hostage ten customers and eight employees of a Lindt chocolate caf é located at Martin Place in Sydney, Australia, on December 15, 2014.
 \item Charlie Hebdo shooting: two brothers forced their way into the offices of the French satirical weekly newspaper Charlie Hebdo in Paris, killing 11 people and wounding 11 more, on January 7, 2015.
 \item Germanwings plane crash: a passenger plane from Barcelona to Düsseldorf crashed in the French Alps on March 24, 2015, killing all passengers and crew. The plane was ultimately found to have been deliberately crashed by the co-pilot of the plane.
\end{itemize}

Given the large volume tweets in the datasets, we sampled them by picking tweets that provoked a high number of retweets. The retweet threshold was set to 100, selected based on the size of the resulting dataset. For each of these tweets in the sampled subset, we also collect all the tweets that reply to them. While Twitter does not provide an API endpoint to retrieve conversations provoked by tweets, it is possible to collect them by scraping tweets through the web client interface. We developed a script that enabled us to collect and store complete conversations for all the rumourous source tweets\footnote{The conversation collection script is available at https://github.com/azubiaga/pheme-twitter-conversation-collection.}. We use the replying tweets for two purposes: (1) for the manual annotation work, where replies to each tweet can provide context for the annotator where needed to decide if a tweet is a rumour, and (2) we use them to reproduce a baseline classifier based on the baseline introduced by \cite{zhao2015enquiring}. However, our approach ignores replying tweets, relying only on the source tweet itself.

\begin{figure}[ht]
 \centering
 \includegraphics[width=\textwidth]{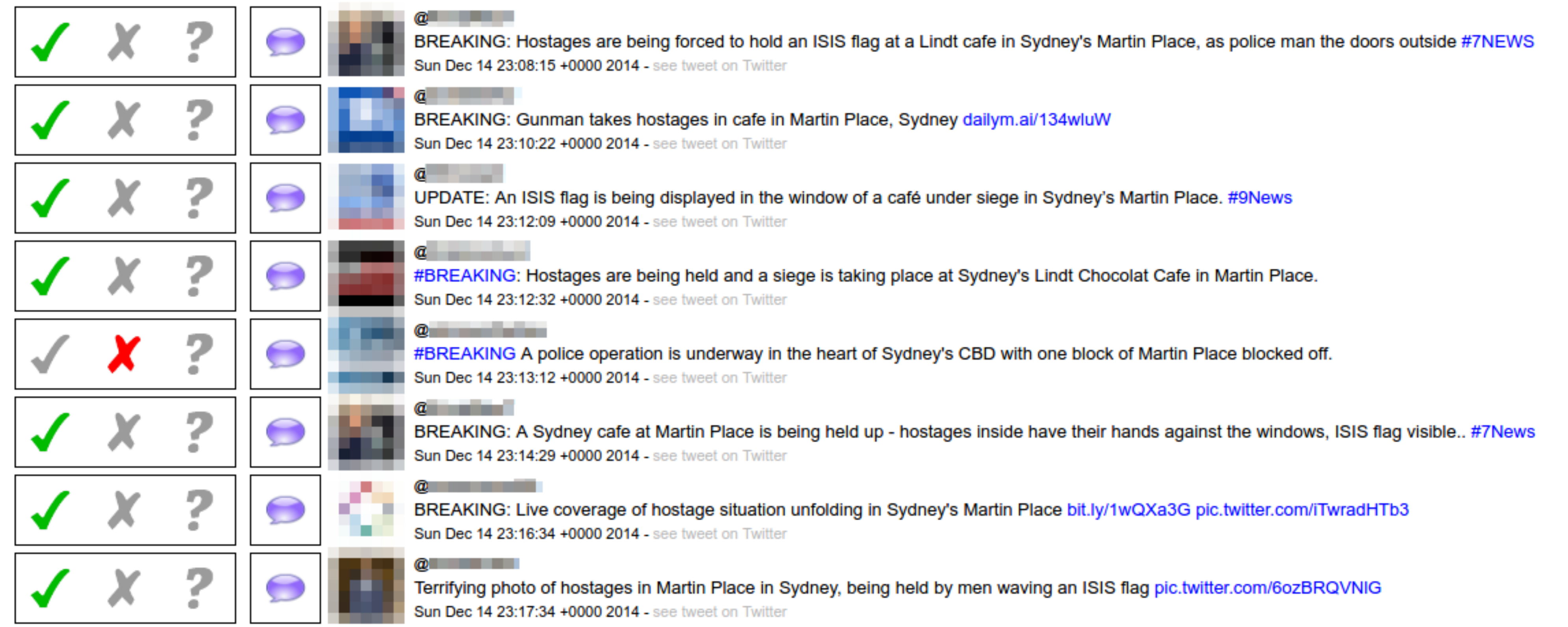}
 \caption{Screenshot of tool that the journalists utilised to annotate tweets in a timeline as being rumours or non-rumours. Each tweet can be annotated as a rumour (green tick) or a non-rumour (red cross). The question mark was solely used to leave a tweet for later. Additionally, the bubble next to each tweet enables to visualise the tweets replying to that tweet, used occasionally for context.}
 \label{fig:annotation-tool}
\end{figure}

The sampled subsets of tweets were visualised in a separate timeline per day and sorted by time (see Figure \ref{fig:annotation-tool}). Using these timelines, journalists were asked to identify rumours and non-rumours. Along with each tweet, journalists could optionally click on the bubble next to the tweet to visualise tweets replying to the tweet; the conversation provoked by the tweet could assist them by providing context, albeit the annotation was independent of this context and based on their experience. The fact that this annotation work was performed by journalists was convenient as they had continually tracked the five events while they were unfolding, and so they were knowledgeable about the stories. The annotation, however, was conducted \textit{a posteriori}, once the reporting about the event had come to an end. This encouraged careful annotation that encompassed a broad set of rumours; the journalists could go through the whole timeline of tweets as we presented them and perform the annotations. The annotation work led to the manual categorisation of each tweet as being a rumour or not. The methodology they followed to perform these annotations in a more manageable and scalable way was to go through the timeline by analysing carefully those tweets that reported new stories that they had not seen before; for those cases, they investigated the story further on social media and the Web when the origin and nature of the story was not known to them. As they progressed in the timeline, new tweets reporting repeated stories where assigned the same annotation as in the previous instance. This made their job easier as they had had only to investigate carefully stories that they had not seen.

\begin{figure}[ht]
 \centering
 \includegraphics[width=\textwidth]{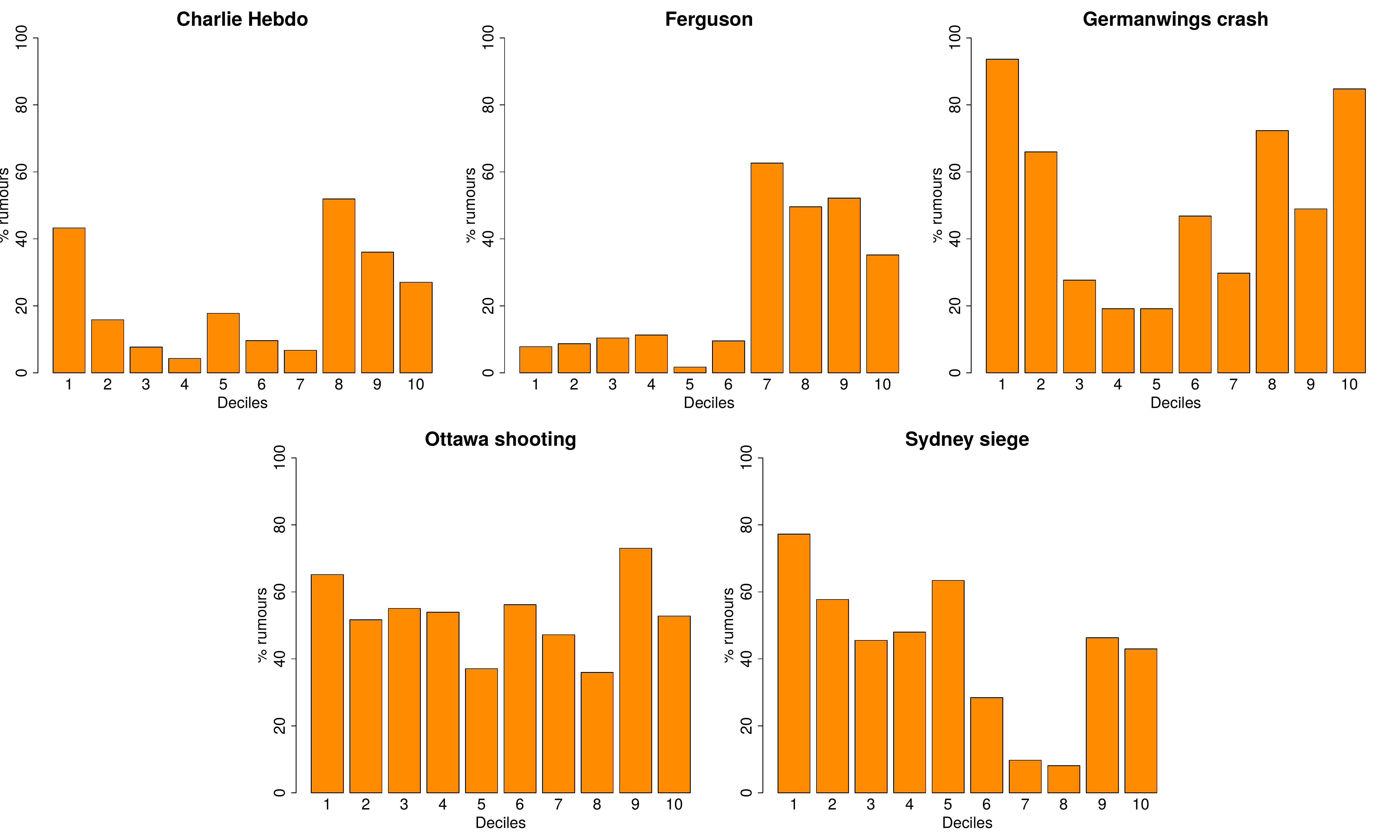}
 \caption{Rumour ratios for deciles within the timeline of each event, showing how ratios of rumours evolve as the event unfolds.}
 \label{fig:deciles}
\end{figure}

The annotation of tweets sampled for all five events led to a collection of 5,802 annotated tweets, of which 1,972 were deemed rumours and 3,830 were deemed non-rumours. These annotations are distributed differently across the five events, as shown in Table \ref{tab:annotation-distribution}. While slightly over 50\% of the tweets were rumours for the Germanwings Crash and the Ottawa Shooting, less than 25\% were so for Charlie Hebdo and Ferguson. The Sydney Siege had an intermediate ratio of rumours (42.8\%). While the global figures of rumours vary substantially across events, we dug further into these distributions to understand how rumours and non-rumours are distributed during events, e.g., to look at whether rumours occur especially at the beginning of the event, along with the very early reports. To do this, we broke down the timeline of tweets for each event into deciles (10\% percentiles) and look at the ratio of rumours in each of these deciles. Figure \ref{fig:deciles} shows the ratios of rumours for each of the deciles for the five events in our dataset. Contrary to what we initially expected, there is no common pattern across events. One can see events with uniformly distributed ratios of rumours, such as with the Ottawa Shooting, events where the ratio of rumours fades at least eventually, such as Charlie Hebdo, Germanwings crash and Sydney Siege, or events where the majority of the rumours emerge in later stages of the reporting, such as Ferguson. These varying distributions of rumours across different events makes the rumour detection task even more challenging, as one may not be able to rely on the earliness of a report to determine if it is more likely to be a rumour.

\begin{table*}[ht]
 \centering
 \caption{Distribution of annotations of rumours and non-rumours for the five breaking news in the dataset.}
 \begin{tabular}{l r r r}
  \toprule
  Event & Rumours & Non-rumours & Total \\
  \midrule
  Charlie Hebdo & 458 (22.0\%) & 1,621 (78.0\%) & 2,079 \\
  Ferguson & 284 (24.8\%) & 859 (75.2\%) & 1,143 \\
  Germanwings Crash & 238 (50.7\%) & 231 (49.3\%) & 469 \\
  Ottawa Shooting & 470 (52.8\%) & 420 (47.2\%) & 890 \\
  Sydney Siege & 522 (42.8\%) & 699 (57.2\%) & 1,221 \\
  \midrule
  Total & 1,972 (34.0\%) & 3,830 (66.0\%) & 5,802 \\
  \bottomrule
 \end{tabular}
 \label{tab:annotation-distribution}
\end{table*}

\section{Rumour Detection Task}

We define the rumour detection task as that in which, given a timeline of tweets, the system has to determine which of the tweets are reporting rumours, and hence are spreading information that is yet to be verified. The identification of rumours within a timeline is ultimately meant to warn users that the information has not been confirmed, and so it may turn out to be false. This can be operationalised by flagging those tweets that are identified as rumours, warning users to think twice before spreading the information. Formally, the task takes an evolving timeline of tweets $TL = \{t_1,..., t_{|TL|}\}$ as input, and the classifier has to determine whether each of these tweets, $t_i$, is a rumour or a non-rumour by assigning a label from $Y = \{R, NR\}$.

Hence, we formulate the task as a binary classification problem, whose performance is evaluated by computing the precision, recall and F1 scores for the target category, i.e., rumours.

\section{Learning Sequential Dynamics for Rumour Detection}

\subsection{Hypothesis}

We argue that a single headline or tweet may not always be indicative of a piece of information being a rumour. There are, indeed, cases where a single tweet uses hedging words or provides little or no evidence so as to be deemed corroborated information, and hence those cases can be deemed rumours from the tweet alone. This is the case, for instance, of tweets reporting during the Ferguson riots that ``\textit{the name of the police officer who fatally shot the kid would be reportedly announced by the police later in the day}''. If the tweet itself expresses uncertainty, as is the case here with the use of ``reportedly'', one can consider that the underlying information is not confirmed. However, reports confidently reporting that ``\textit{the kid was involved in a robbery before being shot}'' may not be as easily identified by an automated classifier from the tweet alone, despite being a rumour. The dearth of sufficient evidence as occurs in many tweets encourages us to further leverage context that could help the classifier distinguish rumours and non-rumours.

One possibility to extend a tweet with context is to look at how others react to it, as \cite{zhao2015enquiring} proposed in their work that querying or enquiring tweets provoked by a posting may indicate it is a rumour. However, we believe the public will not always question the veracity of rumours, given that average users may not always notice that a piece of information is not confirmed. This is the case of a number of tweets during the Ferguson riots reporting that ``\textit{the kid was shot 10 times by the police}''. While this information was not queried by the public, the media treated the information as not being verified; the autopsy later confirmed that he was shot 6 times. Hence, while reactions may be indicative of a piece of information being unverified, we believe that it may lead to low recall, missing other cases that are not rebutted.

To better harness the context surrounding a tweet, we believe that the classifier needs to be aware of how the whole event is unfolding, analysing the different announcements that build a story before the current tweet is posted. The tweet that is being classified as rumour or non-rumour should therefore leverage earlier happenings within that event, both rumours and non-rumours, that make up a story in which the current tweet fits. For instance, a tweet reporting the rumour that ``\textit{the police officer who shot the kid has left the town}'' may be easier to classify being aware of all the previous reports related to the police officer and the killing. Based on this, we set forth the hypothesis that \textit{aggregation of all the rumourous and non-rumourous reports leading up to the tweet being classified will provide key context to boost performance of the rumour detection system}.

\subsection{Classifiers}

In order to test our hypothesis, we use Conditional Random Fields (CRF) as a sequential classifier that enables aggregation of tweets as a chain of reports. We use a Maximum Entropy classifier as the non-sequential equivalent of CRF to test the validity of the hypothesis, and also use additional baseline classifiers for further comparison. Moreover, we also reproduce a baseline based on the approach introduced by \cite{zhao2015enquiring} to compare the performance of our approach with that of a state-of-the-art approach.

\textbf{Conditional Random Fields (CRF).} We use CRF as a structured classifier to model sequences of tweets as observed in the timelines of Twitter breaking news. With CRF, we can model the timeline as a linear chain or graph that will be treated as a sequence of rumours and non-rumours. In contrast to classifiers traditionally used for this task, which choose a label for each input unit (e.g., a tweet), CRF also consider the neighbours of each unit, learning the probabilities of transitions of label pairs to be followed by each other. The input for CRF is a graph $G = (V, E)$, where in our case each of the vertices $V$ is a tweet, and the edges $E$ are relations of tweets, i.e., a link between a tweet and its preceding tweet in the event. Hence, having a data sequence $X$ as input, CRF outputs a sequence of labels $Y$ \cite{lafferty2001conditional}, where the output of each element $y_i$ will not only depend on its features, but also on the probabilities of other labels surrounding it. The generalisable conditional distribution of CRF is shown in Equation \ref{eq:crf} \cite{sutton2011introduction}\footnote{We use the PyStruct to implement Conditional Random Fields \cite{muller2014pystruct}.}.

\begin{equation}
 p(y|x) = \frac{1}{Z(x)} \prod_{a = 1}^{A} \Psi_a (y_a, x_a)
 \label{eq:crf}
\end{equation}

where Z(x) is the normalisation constant, and $\Psi_a$ is the set of factors in the graph $G$.

Therefore, in our specific case of rumour detection, CRF will exploit the sequence of rumours and non-rumours leading up to the current tweet to determine if it is a rumour or not. It is important to note that with CRF the sequence of rumours and non-rumours preceding the tweet being classified will be based on the predictions of the classifier itself, and will not use any ground truth annotations. Errors in early tweets in the sequence may then augment errors in subsequent tweets. For each tweet to be classified, we solely feed the preceding tweets to the classifier to simulate a realistic scenario where subsequent tweets are not yet posted and early decisions need to be made on each tweet.

\textbf{Maximum Entropy classifier (MaxEnt).} As the non-sequential equivalent of CRF, we use a Maximum Entropy (or logistic regression) classifier, which is also a conditional classifier but which will operate at the tweet level, ignoring the sequence and hence the preceding tweets. This enables us to compare directly the extent to which treating the tweets posted during an event as a sequence instead of having each tweet as a separate unit can boost the performance of the classifier.

\textbf{Enquiry-based approach by \cite{zhao2015enquiring}:} As a state-of-the-art baseline for rumour detection, and the only approach that so far has tackled rumour detection in social media, we reproduce the approach by Zhao et al., which uses regular expressions to look for enquiry posts. We use the set of replies responding to each tweet to look for enquiry posts. Following the approach described by the authors, we consider that a tweet is a rumour if at least one of the replying tweets matches with one of the regular expressions that the authors curated. The list of regular expressions defined by the authors is shown in Table \ref{tab:www2015-regex}.

\begin{table}[ht]
 \centering
 \caption{List of regular expressions utilised by Zhao et al., which we reimplemented to reproduce their approach as a baseline. Regular expressions for both enquires and corrections are combined, and a tweet that matches any of them will be deemed an enquiry tweet.}
 \begin{tabular}{l l}
  \toprule
  Pattern Regular Expression & Type \\
  \midrule
  is (that \textbar~this \textbar~it) true & Verification \\
  wh[a]*t[?!][?1]* & Verification \\
  ( real? \textbar~really ? \textbar~unconfirmed )  & Verification \\
  (rumor \textbar~debunk) & Correction \\
  (that \textbar~this \textbar~it) is not true & Correction \\
  \bottomrule
 \end{tabular}
 \label{tab:www2015-regex}
\end{table}

\textbf{Additional baselines.} We also compare three more non-sequential classifiers\footnote{We use their implementation in the scikit-learn Python package for Maximum Entropy, Naive Bayes, Support Vector Machines and Random Forests.}: Naive Bayes (NB), Support Vector Machines (SVM), and Random Forests (RF).

We perform the experiments in a 5-fold cross-validation setting, having in each case four of the events for training, and the remainder event for testing. This enables us to simulate a realistic scenario where an event is completely unknown to the classifier and it has to identify rumours from the knowledge garnered from events in the training set. For evaluation purposes, we aggregate the output of all five runs as the micro-averaged evaluation across runs.

\subsection{Features}

We use two types of features with the classifiers: content-based features and social features. We test them separately as well as combined. The features that fall in each of these categories are as follows:

\subsubsection{Content-based Features}

We use seven different features extracted from the content of the tweets:

\begin{itemize}
 \item \textbf{Word Vectors:} to create vectors representing the words in each tweet, we build word vector representations using Word2Vec \cite{mikolov2013distributed}. We train a different Word2Vec model with 300 dimensions for each of the five folds, training the model in each case from the collection of tweets pertaining to the four events in the training set, so that the event (and the vocabulary) in the test set is unknown. As a result, we get five different Word2Vec models, each used in a separate fold.
 \item \textbf{Part-of-speech Tags:} we build a vector of part-of-speech (POS) tags with each feature in the vector representing the number of occurrences of a certain POS tag in the tweet. We use Twitie \cite{bontcheva2013twitie} to parse the tweets for POS tags, an information extraction package that is part of GATE \cite{cunningham2011text}.
 \item \textbf{Capital Ratio:} the ratio of capital letters among all alphabetic characters in the tweet. Use of capitalisation tends to represent emphasis, among others.
 \item \textbf{Word Count:} the number of words in the tweet, counted as the number of space-separated tokens.
 \item \textbf{Use of Question Mark:} a binary feature representing if the tweet has at least a question mark in it. Question marks may be indicative of uncertainty.
 \item \textbf{Use of Exclamation Mark:} a binary feature representing if the tweet has at least an exclamation mark in it. Exclamation marks may be indicative of emphasis or surprise.
 \item \textbf{Use of Period:} a binary feature representing if the tweet has at least a period in it. Punctuation may be indicative of good writing and hence potentially of slow reporting.
\end{itemize}

\subsubsection{Social Features}

We use five social features, all of which can be inferred from the metadata associated with the author of the tweet, and which is embedded as part of a tweet object retrieved from the Twitter API. We define a set of social features that are indicative of a user's experience and reputation:

\begin{itemize}
 \item \textbf{Tweet Count:} we infer this feature from the number of tweets that a user has posted on Twitter. As numbers can vary substantially across users, we normalise them by rounding up the 10-base logarithm of the tweet count: $\left \lceil{\log_{10}(\text{statusescount})}\right \rceil$.
 \item \textbf{Listed Count:} this feature is computed by normalising the number of lists a user belongs to, i.e., the number of times other users decided to add them to a list: $\left \lceil{\log_{10}(\text{listedcount})}\right \rceil$.
 \item \textbf{Follow Ratio:} in this feature we look at the reputation of a user as reflected by their number of followers. However, the number of followers might occasionally be rigged, e.g., by users who simply follow many others to attract more followers. To control for this effect, we define the follow ratio as the logarithmically scaled ratio of followers over followees: $\left \lfloor{\log_{10}~(\text{\#followers}/\text{\#following})}\right \rceil$.
 \item \textbf{Age:} we compute the age of a user as the rounded number of years that the user has spent on Twitter, i.e., from the day the account was set up to the day of the current tweet.
 \item \textbf{Verified:} a binary feature representing if the user has been verified by Twitter or not. Verified users are those whose identity Twitter has validated, and tend to be reputable people.
\end{itemize}

\section{Results}

\subsection{Comparison of Classifiers}

Table \ref{tab:comparison-classifiers} shows the results for different classifiers using either or both of the content-based and social features, as well as the results for the state-of-the-art classifier by \cite{zhao2015enquiring}. Performance results of the classifiers using content-based features suggests a remarkable improvement for CRF over the rest of the classifiers, implying that CRF benefits from the use of the sequence of tweets preceding each tweet as context to enrich the input to the classifier. This is especially true when we look at  precision, where CRF performs substantially better than the rest. Only the Naive Bayes classifier performs better in terms of recall, however, it performs poorly in terms of precision. As a result, CRF balances precision and recall in a clearly better way, outperforming all the other classifiers in terms of the F1 score.

Results are not as clear when we look at those using social features. CRF still performs best in terms of precision, but performance drops if we look at the recall. In fact, most of the classifiers perform better than CRF in terms of recall, with SVM as the best performing classifier. Combining both precision and recall in an F1 score shows that SVM is the classifier that best exploits social features. However, performance results using social features are significantly worse than those using content-based features, which suggests that social features alone are not sufficient.

When both content-based features and social features are combined as an input to the classifier, we see that the results resemble that of the use of content-based features alone. CRF outperforms all the rest in terms of precision, while Naive Bayes is good only in terms of recall. As a result, the aggregation of features also leads to CRF being the best classifier in terms of F1 score. In fact, CRF leads to an improvement of 39.9\% over the second best classifier in terms of F1, Naive Bayes. If we compare the results of CRF with the use of content-based features alone or combining both types of features, we notice that the improvement comes especially for recall, which is balanced out with a slight drop of precision. As a result, we get an F1 score that is slightly better when using both features together. In fact, all F1 scores for combined features are superior to their counterparts using content-based features alone, among which CRF performs best.

Comparison with respect to the enquiry-based baseline approach introduced by Zhao et al. buttresses our conjecture that a manually curated list of regular expressions may lead to low recall, which is as low as 0.065 in this case. This approach gets a relatively good precision score, which beats all of our baselines, although it performs substantially worse than CRF. However, 59\% of false positives as can be inferred from the precision of 0.41 indicates that the regular expressions also match non-rumours. One could also opt to expand the list of regular expressions and/or adapt them to our specific scenario and events; however, this may involve substantial manual work and would not guarantee generalisable performance.

\begin{table}[ht]
 \centering
 \caption{Results}
 
 \begin{tabular}{l r r r}
  \toprule
  \multicolumn{4}{c}{Content} \\
  \midrule
  Classifier & P & R & F1 \\
  \midrule
  SVM & 0.355 & 0.445 & 0.395 \\
  Random Forest & 0.271 & 0.087 & 0.131 \\
  Naive Bayes & 0.309 & \textbf{0.723} & 0.433 \\
  Maximum Entropy & 0.329 & 0.425 & 0.371 \\
  CRF & \textbf{0.683} & 0.545 & \textbf{0.606} \\
  \bottomrule
 \end{tabular}
 
 \begin{tabular}{l r r r}
  \toprule
  \multicolumn{4}{c}{Social} \\
  \midrule
  Classifier & P & R & F1 \\
  \midrule
  SVM & 0.337 & \textbf{0.524} & \textbf{0.410} \\
  Random Forest & 0.343 & 0.433 & 0.382 \\
  Naive Bayes & 0.294 & 0.010 & 0.020 \\
  Maximum Entropy & 0.336 & 0.476 & 0.394 \\
  CRF & \textbf{0.462} & 0.268 & 0.339 \\
  \bottomrule
 \end{tabular}
 
 \begin{tabular}{l r r r}
  \toprule
  \multicolumn{4}{c}{Content + Social} \\
  \midrule
  Classifier & P & R & F1 \\
  \midrule
  SVM & 0.337 & 0.483 & 0.397 \\
  Random Forest & 0.275 & 0.099 & 0.145 \\
  Naive Bayes & 0.310 & \textbf{0.723} & 0.434 \\
  Maximum Entropy & 0.338 & 0.442 & 0.383 \\
  CRF & \textbf{0.667} & 0.556 & \textbf{0.607} \\
  \bottomrule
 \end{tabular}
 \begin{tabular}{l r r r}
  \toprule
  \multicolumn{4}{c}{State-of-the-art Baseline} \\
  \midrule
  Classifier & P & R & F1 \\
  \midrule
  \cite{zhao2015enquiring} & 0.410 & 0.065 & 0.113 \\
  \bottomrule
 \end{tabular}
 
 \label{tab:comparison-classifiers}
\end{table}

\subsection{Consistency of the Sequential Classifier's Performance}

\begin{figure}[ht]
 \centering
 \includegraphics[width=\textwidth]{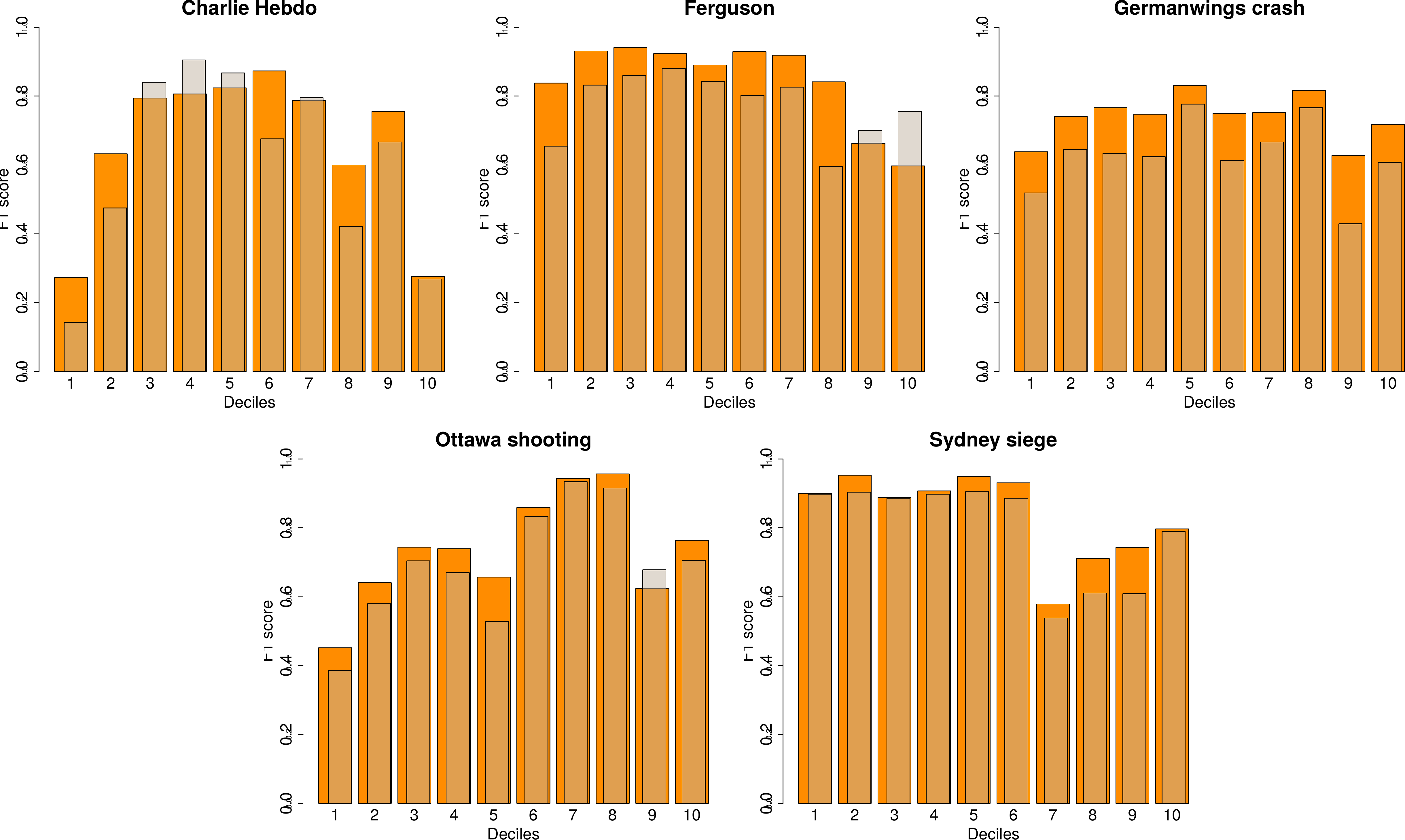}
 \caption{F1 scores by event and decile using CRF (orange, thick bars) vs the best of the non-sequential classifiers in each decile (grey, thin bars).}
 \label{fig:decile-f1s}
\end{figure}

Even though CRF as a sequential classifier has proven to perform better than the rest of the non-sequential classifiers overall, we are interested in seeing whether the high performance of CRF is consistent over time. Given that CRF depends on the sequence as context to enhance classification of a tweet, the first few tweets in each event lack the context that later tweets have. To analyse performance over time, we look at the F1 scores for each decile and for each event separately. Figure \ref{fig:decile-f1s} shows these results, broken down by event and decile; thick, orange bars represent the F1 score of the CRF classifier in each decile, while the thinner, grey bars represent the highest F1 score across all non-sequential classifiers in each decile. We make some interesting observations from these results:

\begin{itemize}
 \item CRF generally outperforms the best of the non-sequential classifiers, also when breaking down the results by decile; for the 50 deciles in our dataset, this is true in 43 of the cases, with only 7 cases where another classifier performs better in a decile.
 \item CRF does suffer from a cold start problem at the beginning of each event, which is especially noticeable with Charlie Hebdo and Ottawa shootings. In the five events under study, CRF performs better in the second decile than in the first, and it tends to perform better in later deciles than in the first. This may indicate, as we conjectured, that CRF can only make use of a short sequence, providing little context, when classifying the very first tweets. Nevertheless, CRF also shows better performance for the initial deciles than the best of the non-sequential classifiers.
 \item Not all the events have the same characteristics when it comes to rumour spawning, and for some events it even becomes challenging to detect rumours in later stages. This occurs, for instance, with Charlie Hebdo, where performance drops in the 10th decile, or with Sydney siege, where performance drops in the 7th decile and then it progressively recovers again. What is interesting is that both CRF and the non-sequential classifiers show similar increasing and decreasing trends across events and deciles, and that it is the event -- rather than the classifier -- which makes certain stages more challenging.
\end{itemize}

\section{Discussion}

Our rumour detection experiments on five datasets, each associated with a breaking news story, show that a classifier that sequentially exploits context from earlier tweets achieves significant improvements over standard, non-sequential classifiers. We have proven this in the case of a CRF classifier, which substantially outperforms its non-sequential counterpart, a Maximum Entropy classifier, as well as other non-sequential classifiers. Moreover, our approach also beats the state-of-the-art baseline by \cite{zhao2015enquiring} that uses regular expressions to classify as rumours the tweets that provoke reactions matching certain patterns. The latter fails to achieve a competitive recall score, which we argue is for two main reasons: (1) rumours will not always ignite enquiring reactions from the general public, and (2) the manually curated regular expressions might be limited, requiring regular updates involving a human in the loop. Our fully automated sequential classifiers can, instead, classify a tweet as a rumour or non-rumour from its own content and context from earlier tweets, without having to wait for any reactions.

In contrast with previous work in related tasks dealing with rumours, our work here has covered a wide range of rumours and non-rumours that were not necessarily known \textit{a priori}. This was possible thanks to having as annotators a team of journalists who had followed the events closely and for the way the annotation work was performed, i.e., showing them a timeline of tweets that enabled them discovering rumours and non-rumours that one may have initially missed. While we are confident that this approach covers a diverse range of rumours and non-rumours, one caveat that it is important to note is that it is restricted to a subset of highly retweeted tweets. Consequently, our experiments have been limited to tweets being retweeted at least 100 times. This is consistent with one of the key characteristics of rumours, i.e., that they have to attract a substantial interest to be deemed rumours. While this is sensible for the task of rumour detection and the objectives of our work, it is necessary to wait until a tweet gets retweeted a number of times before it can be considered a candidate for input to the classifier. The development of a classifier that identifies these highly retweeted tweets promptly would enable early detection of rumours by not having to wait for the tweets to reach a certain threshold of retweets.

\section{Conclusion}

We have introduced a novel approach to rumour detection in social media by leveraging the context preceding a tweet with a sequential classifier. Experimenting over five breaking news datasets collected from Twitter and annotated for rumours and non-rumours by journalists, we show that the preceding context exploited as a sequence can substantially boost the rumour detector's performance. Our approach has also proven to outperform the state-of-the-art rumour detection system introduced by \cite{zhao2015enquiring} that, instead, relies on find querying posts that match a set of manually curated list of regular expressions. Their approach performs well in terms of precision but fails in terms of recall, suggesting that regular manual input is needed to revise the regular expressions. Our fully automated approach instead achieves superior performance that is better balanced for both precision and recall.

Social media and user-generated content (UGC) are increasingly important features in a number of different ways for the work of not only journalists but also government agencies such as the police and civil protection agencies \cite{procter2013readingb}. However, their use present major challenges, not least because information posted on social media is not always reliable and its veracity needs to be checked before it can be considered as fit for use in the reporting of news, or decision-making in the case of criminal activity \cite{procter2013readingb} or disaster response \cite{bazerli2015humanitarianism}. Hence, it is vital that tools be developed that can aid a) the detection of rumours and b) determining their likely veracity. In the Pheme project \cite{derczynski2015pheme}, we have been developing tools that address the need for the latter \cite{zubiaga2016analysing,lukasik2016using}. However, for tools for rumour veracity determination to be effective, they need to be applied in combination with the former and progress so far has been limited. In this paper, we present a novel approach whose performance suggests it has the potential to address the former problem. 

With this work we also make the annotated datasets publicly available to enable and encourage further research in the task\footnote{\url{https://figshare.com/articles/PHEME_dataset_of_rumours_and_non-rumours/4010619}}.

\section{Acknowledgments}

This work has been supported by the PHEME FP7 project (grant No. 611233). This research utilised Queen Mary's MidPlus computational facilities, supported by QMUL Research-IT and funded by EPSRC grant EP/K000128/1.

\bibliographystyle{apalike}
\bibliography{rnr}

\label{lastpage}

\end{document}